\pdfoutput=1

\documentclass[letterpaper, 10 pt, conference]{ieeeconf}  

\IEEEoverridecommandlockouts                              

\usepackage{graphicx}
\usepackage{algorithm}   
\usepackage{algorithmic}
\usepackage{wrapfig}
\usepackage{amsfonts}
\usepackage{booktabs}
\usepackage{multirow}
\usepackage{pifont}
\usepackage[table]{xcolor}
\usepackage[most]{tcolorbox}
\usepackage{mathrsfs}
\definecolor{LightGray}{gray}{0.97}
\definecolor{LightBlue}{rgb}{0.97,0.985,1.0}
\definecolor{LightOrange}{rgb}{1.0,0.97,0.94}
\definecolor{LightGreen}{rgb}{0.98,1.0,0.98}
\definecolor{LightPurple}{rgb}{0.96,1.0,1.0}

\usepackage{threeparttable}
\usepackage{amsmath,amssymb}
\usepackage{hyperref}
\usepackage{xspace}
\DeclareRobustCommand{\vlnbert}{%
  VLN\texorpdfstring{\ensuremath{\mathbin{\circlearrowright}}}{->}BERT\xspace}

\title{\LARGE \bf
StreamVLN: Streaming Vision-and-Language Navigation \\via SlowFast Context Modeling
}

\author{Meng Wei$^{*1,2}$, Chenyang Wan$^{*1,3}$, Xiqian Yu$^{*1}$, Tai Wang$^{*1,\dagger}$, Yuqiang Yang$^{1}$, Xiaohan Mao$^{1,4}$, \\Chenming Zhu$^{1,2}$, Wenzhe Cai$^{1}$, Hanqing Wang$^{1}$, Yilun Chen$^{1}$,Xihui Liu$^{2,\ddagger}$,Jiangmiao Pang$^{1,\ddagger}$
\thanks{$^{1}$Shanghai AI Lab, $^{2}$The University of Hong Kong, $^{3}$Zhejiang University $^{4}$Shanghai Jiao Tong University}\thanks{$^{*}$Equal Contribution, $^{\dagger}$Project Lead, $^{\ddagger}$Corresponding Authors}}

\begin{document}

\maketitle
\thispagestyle{empty}
\pagestyle{empty}

\begin{abstract}
    Vision-and-Language Navigation (VLN) in real-world settings requires agents to process continuous visual streams and generate actions with low latency grounded in language instructions. While Video-based Large Language Models (Video-LLMs) have driven recent progress, current VLN methods based on Video-LLM often face trade-offs among fine-grained visual understanding, long-term context modeling and computational efficiency.  
    We introduce StreamVLN, a streaming VLN framework that employs a hybrid slow-fast context modeling strategy to support multi-modal reasoning over interleaved vision, language and action inputs.
    The fast-streaming dialogue context facilitates responsive action generation through a sliding-window of multi-turn dialogues, while the slow-updating memory context compresses historical visual states using a 3D-aware token pruning strategy.
    With this slow-fast design, StreamVLN achieves real-time dialogues through KV cache reuse, supporting long video streams with bounded context size and inference cost.
    Experiments on VLN-CE benchmarks show state-of-the-art performance with low latency, ensuring robustness and efficiency in real-world deployment. 
    The project page is: \href{https://streamvln.github.io/}{https://streamvln.github.io/}.
\end{abstract} 
\section{Introduction}
\label{sec:intro}
Vision-and-Language Navigation (VLN) in continuous real-world environments is a critical task in embodied AI, where an agent must 
ground linguistic cues in visual observations and plan actionable trajectories.
However, achieving robust VLN remains challenging due to the need for fine-grained multimodal alignment, long-term sequence reasoning, and generalization to unseen environments.
Recent advances in Video-LLMs offer new capabilities for VLN systems.
Several research efforts~\cite{zhang2024navid,zhang2024uninavid,cheng2024navila} have extended Video-LLMs to vision-language-action models (VLA) for navigation, which integrate visual observation encoding, language understanding, and action prediction in a unified end-to-end framework.

For real-world navigation, VLA models must process continuously incoming video streams, where maintaining long-term context and real-time responsiveness are both crucial. This poses challenges for Video-LLMs in managing linearly growing visual tokens.
Some methods~\cite{cheng2024navila,zhang2025mapnav} sample a fixed number of video frames, but the limited temporal resolution may fail to accurately predict low-level actions when fine-grained temporal changes are needed.
Other methods~\cite{zhang2024navid,zhang2024uninavid} compress vision tokens into sparse memory tokens via pooling or token merging, which helps control the token volume but sacrificing temporal and visual details.
Furthermore, these methods typically require refreshing the LLM’s dialogue context at every action step.
This leads to significant redundant computation during both training and inference, hindering data scalability and real-world deployment.
\begin{figure}
    \centering
    \includegraphics[width=\linewidth]{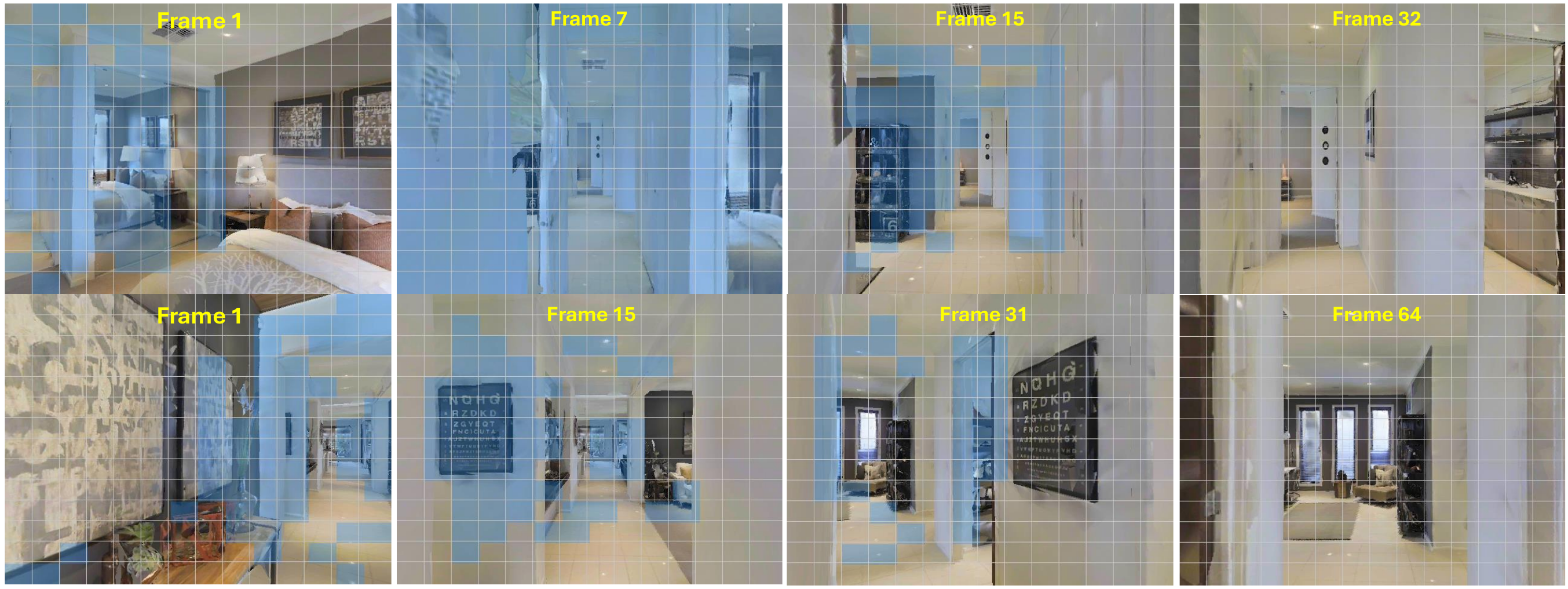}
    \caption{Visualization of the proposed 3D spatial pruning strategy. The spatial-temporal redundancy in ego-centric VLN video data arises from fine-grained low-level action trajectories. While temporal sampling alleviates part of this redundancy, substantial patch-level redundancy remains at the spatial level under high input resolutions of Video-LLMs.}
    \label{fig:voxel_vis}
\end{figure}
\begin{figure*}
    \centering
    \includegraphics[width=\linewidth]{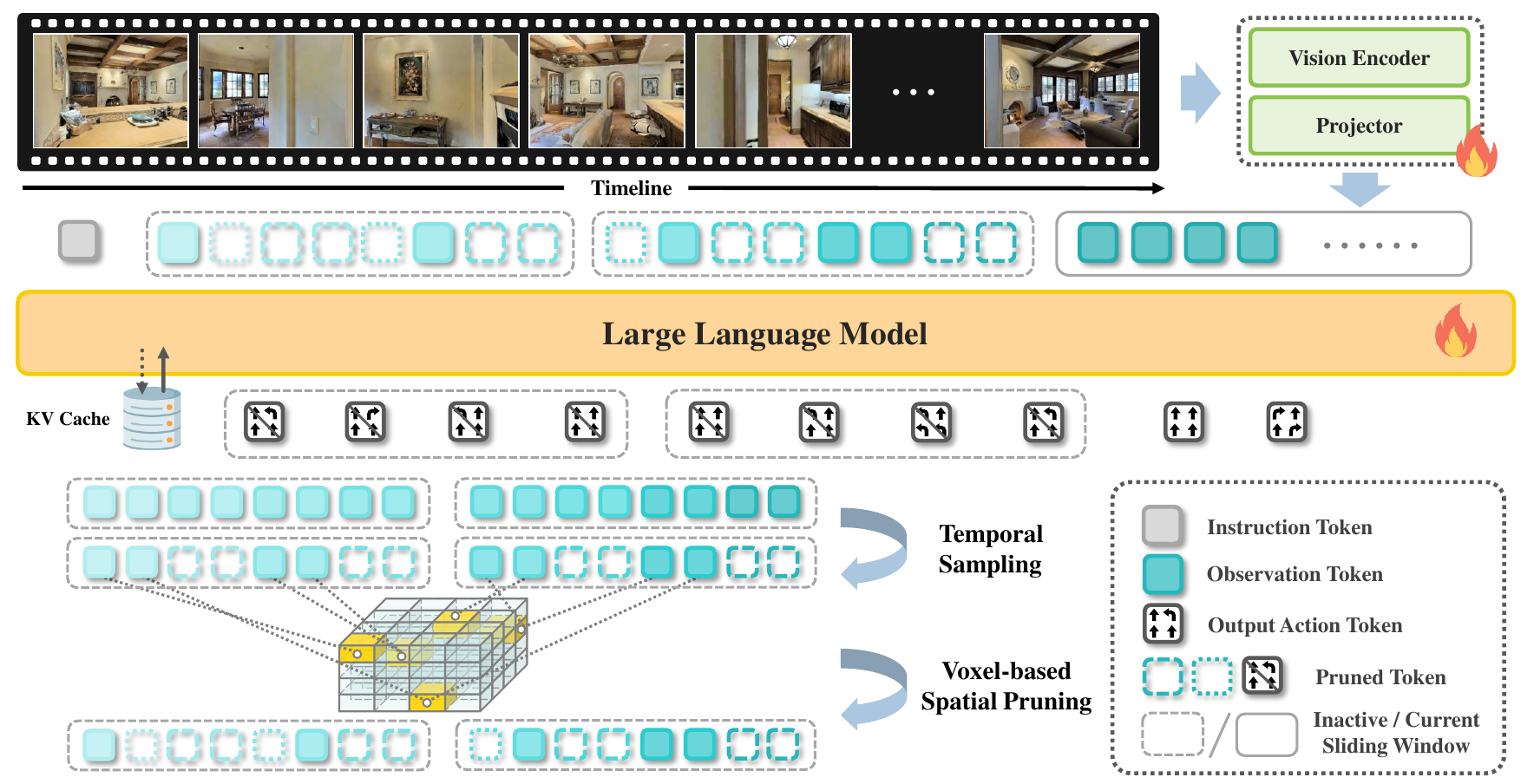}
    \caption{\textbf{Framework of StreamVLN.} The input consists of a language instruction and a stream of RGB images. Each navigation episode is framed as a multi-turn dialogue, where the agent continually queries for the next actions. To support long-horizon reasoning while maintaining a manageable context size and low latency, we adopt a fixed-size sliding window to retain recent dialogue history. The context in inactive windows is updated by token pruning to to reduce memory overhead. }
    \label{fig:overview}
\end{figure*}
In this paper, we propose StreamVLN, a novel streaming vision-and-language navigation framework for low-latency action generation. 
We extend Video-LLM into an interleaved vision-language-action model, enabling continuous interaction with a video stream through multi-turn dialogue.
To address the challenges of long-term context modeling and computational efficiency, StreamVLN introduces a hybrid strategy that combines a \textbf{fast-streaming dialogue context} and a \textbf{slow-updating memory context}.
Specifically, it employs a sliding-window mechanism to cache the key/value states (KV) of tokens over a fixed number of dialogue turns for highly responsive action decoding.
After each streaming dialogue ends, the visual context within the window is consolidated into memory at a slower pace, to provide long-term context for subsequent windows.

Moreover, current state-of-the-art Video-LLMs are trained on high-resolution visual inputs, inevitably generating a large number of visual tokens which cause substantial KV cache and decoding overhead.
However, as shown in Figure~\ref{fig:voxel_vis}, even after temporal sampling, egocentric VLN videos retain high spatial redundancy, where 3D spatial cues offer an effective and efficient means to remove such redundancy.
Hence, we propose a \textbf{training-free} spatial pruning strategy guided by voxel-based 3D proximity.
Compared to video token compression strategies for generic offline videos~\cite{huang2024prunevid}, which often rely on costly token feature similarity computations or operations on large attention matrices, our geometry-based spatial pruning has higher computation efficiency and supports streaming video processing and KV cache compression.

In summary, StreamVLN offers an efficient and scalable solution to suit the continuous interaction requirements of real-time vision-and-language navigation. Its slow-fast context modeling design enables the model trained on short clips (e.g., 16 frames), to work effectively on long video streams, without incurring context length growth or compromising inference latency.
Experiments on existing VLN-CE benchmarks shows that StreamVLN achieves superior performance while maintaining low latency.

\section{Related Work}
\noindent\textbf{Vision-and-Language Navigation (VLN).}
This task requires an agent to follow  language instructions while perceiving and acting in environments. Early progress mainly focused on discrete settings, where agents navigate by ``teleporting'' between predefined nodes of a discrete scene graph~\cite{anderson2018r2r,ku2020rxr,chen2021history,chen2022duet}. This formulation emphasizes high-level decision-making but ignores the challenges of real-world navigation. More recent work~\cite{raychaudhuri2021law,georgakis2022cross,chen2022weakly,an2023etpnav} has focused on continuous environments~\cite{krantz_vlnce_2020}, where agents must perform low-level actions in realistic simulators. To address the increased complexity, some methods incorporate a waypoint predictor~\cite{an2023etpnav,wang2023scaling} pretrained in simulators to propose candidate positions, which are then used to guide high-level navigation decisions. Although these approaches have achieved strong performance, the waypoint predictors typically rely heavily on the training scenes and exhibit limited generalization to unseen scenes. Therefore, more flexible and scalable navigation frameworks is needed to generalize better to long-horizon, real-world setting.

\noindent\textbf{Navigation with Multi-Modal Large Language Models (MLLMs).}
Recent advancements in MLLMs have opened new possibilities for VLN by enabling agents to interpret and reason over natural language instructions in a more generalizable way. Some methods~\cite{long2023discuss,chen20232a2nav,long2024instructnav} directly use LLMs as planner in a training-free manner within a modular framework. But there's still a performance gap compared to task-specific models.
Other lines of work~\cite{zhang2024navid,zhang2024uninavid,cheng2024navila,zhang2025mapnav} further fine-tune Video-based LLMs~\cite{zhang2024llavavideo,lin2024vila,li2024llamavid} to better capture spatial-temporal information and generate low-level actions in an end-to-end manner, but often face challenges in balancing computational efficiency and long-horizon memory retention. StreamVLN aims to better accommodate streaming video input by introducing an efficient and scalable framework that supports action generation with coherent multi-turn reasoning with low-latency response and bounded memory usage.

\section{Method}
StreamVLN generates action outputs from continuous video input in an online, multi-turn dialogue manner. 
Built on LLaVA-Video~\cite{zhang2024llavavideo}, we extend it for interleaved vision, language, and action modeling.
The overall framework of StreamVLN is shown in Figure~\ref{fig:overview}.
We briefly introduce the autoregressive generation in continuous multi-turn dialogues for a streaming VLN process (Section~\ref{ar}).
For both effective context modeling of long sequence and efficient computation for real-time interaction, StreamVLN has:
(1) a fast-streaming dialogue context with a sliding-window KV cache (Section~\ref{short-term}); and
(2) a slow-updating memory via token pruning (Section~\ref{long-term}).
Finally, we describe how we curate the navigation data and incorporate diverse multimodal data for multi-task training (Section~\ref{data}).

\subsection{Preliminary: Multi-Turn Autoregressive Generation}
\label{ar}
A multi-turn dialogue session for VLN consists of a sequence of interleaved observations and actions. In each dialogue $d_i = (o_i, a_i)$, the VLN model receives a new observation $o_i$ and produces an action response $a_i$ conditioned on both the current input and the dialogue history. The full input sequence at step 
$i$ is constructed as: $o_{1}a_{1}o_{2}a_{2}...o_{i-1}a_{i-1}$.
In this streaming setting, new tokens from $o_i$ are appended to the token stream continuously. The response $a_i$ is generated token-by-token via autoregressive decoding.
For each dialogue turn, Transformer-based LLMs first perform a \textbf{prefill phase} to encode input tokens, caching their key/value (KV) states in attention layers. These cached KV pairs are then used in the \textbf{decoding phase} to generate new tokens. If we don't use KV cache across turns, the model will repeat this prefilling process of all previous tokens for a new dialogue.
\subsection{Fast-Streaming Dialogue Context}
\label{short-term}
While multi-turn KV cache reuse can eliminate over $99\%$ of prefilling time, it introduces  substantial memory overhead. As the number of dialogues increases, the KV cache grows linearly (e.g., 2K tokens can consume around 5GB of memory), making long sessions impractical. In addition, existing Video-LLMs tend to exhibit degraded reasoning performance when processing overly long contexts.

To manage dialogue context, we adopt a sliding window KV cache over continuous dialogues, retaining a fixed number \(N\) of recent dialogues in an active window:
$W_j = [o_{(i-N+1)}a_{(i-N+1)}...o_{i}a_{i}]$
When the window reaches capacity, the key/value states are offloaded from the LLM, and the states of non-observation dialogue tokens, such as prompts and generated actions, are immediately discarded. 
For the new sliding window, the token states from past windows are processed into memory token states $\left\{ \mathcal{M}_0, \ldots, \mathcal{M}_j \right\}$ (as detailed in Section~\ref{long-term}).
Formally, for the latest observation \(o_i\), the decoder generates \(a_i\) based on the cached token states and the current window's KV cache:
\[
\begin{aligned}
a_i^{W_{j+1}} = \mathrm{Decoder}\bigl(&o_i, \{\mathcal{M}_0, \ldots, \mathcal{M}_j\}, \\
 &\{k_{(i-N+1)} v_{(i-N+1)}, \ldots, k_{(i-1)} v_{(i-1)}\}\bigr).
\end{aligned}
\]
\subsection{Slow-Updating Memory Context}
\label{long-term}
We observe that most VLN trajectories, collected with fine-grained low-level actions, contain redundant observations. To mitigate this, we first adopt a fixed-number temporal sampling strategy following \cite{cheng2024navila}. However, at the spatial level, heavy patch-level redundancy remains under the high input resolution of Video-LLMs. Direct feature-level compression during training—such as average pooling or learnable Q-Formers—led to substantial performance degradation in our experiments, as it altered the pretrained input distribution, thereby undermining pretraining knowledge.

To address the spatial redundancy without disrupting pretrained features, we introduce a training-free voxel-based 3D spatial pruning strategy, which is also tailored for streaming video processing and KV cache memory.
Specifically, we back-project the 2D image patches from the video stream into a shared 3D space using depth information. By discretizing this 3D space into uniform voxels, we can track the voxel indices of the patch tokens over time. If multiple tokens from different frames within a given duration are projected into the same voxel, only the token from the most recent observation is retained, as detailed in Algorithm~\ref{alg:voxel_pruning}. The voxel pruning mask $M$ is then used to select the preserved token states. 

\begin{algorithm}[H]
\caption{Voxel-Based Spatial Pruning}
\label{alg:voxel_pruning}

\begin{algorithmic}[1]
\STATE Voxel map $V \in \mathbb{Z}^{T \times H \times W}$, stride $K$, threshold $\theta$
\STATE Pruning Mask $M \in \{0,1\}^{T \times H \times W}$

\STATE Initialize $M \leftarrow \mathbf{0}$, map $\texttt{latest} \leftarrow \emptyset$

\FOR{each token $(t, x, y)$ with $V_{t,x,y} \geq 0$}
    \STATE $p \leftarrow \left\lfloor t/K \right\rfloor$, $v \leftarrow V_{t,x,y}$
    \IF{$(p, v)$ not in $\texttt{latest}$ or $t$ is newer}
        \STATE $\texttt{latest}[(p, v)] \leftarrow (t, x, y)$
    \ENDIF
\ENDFOR

\STATE Set $M_{t,x,y} \leftarrow 1$ for all $(t, x, y) \in \texttt{latest}$

\STATE For each $t$, if $\sum_{x,y} M_{t,x,y} < \theta \cdot H \cdot W$, set $M_{t,:,:} \leftarrow 0$

\RETURN $M$
\end{algorithmic}
\end{algorithm}

\subsection{Co-Training with Multi-Source Data.}
\label{data}
\begin{figure}[htbp]
    \centering
    \includegraphics[width=0.29\textwidth]{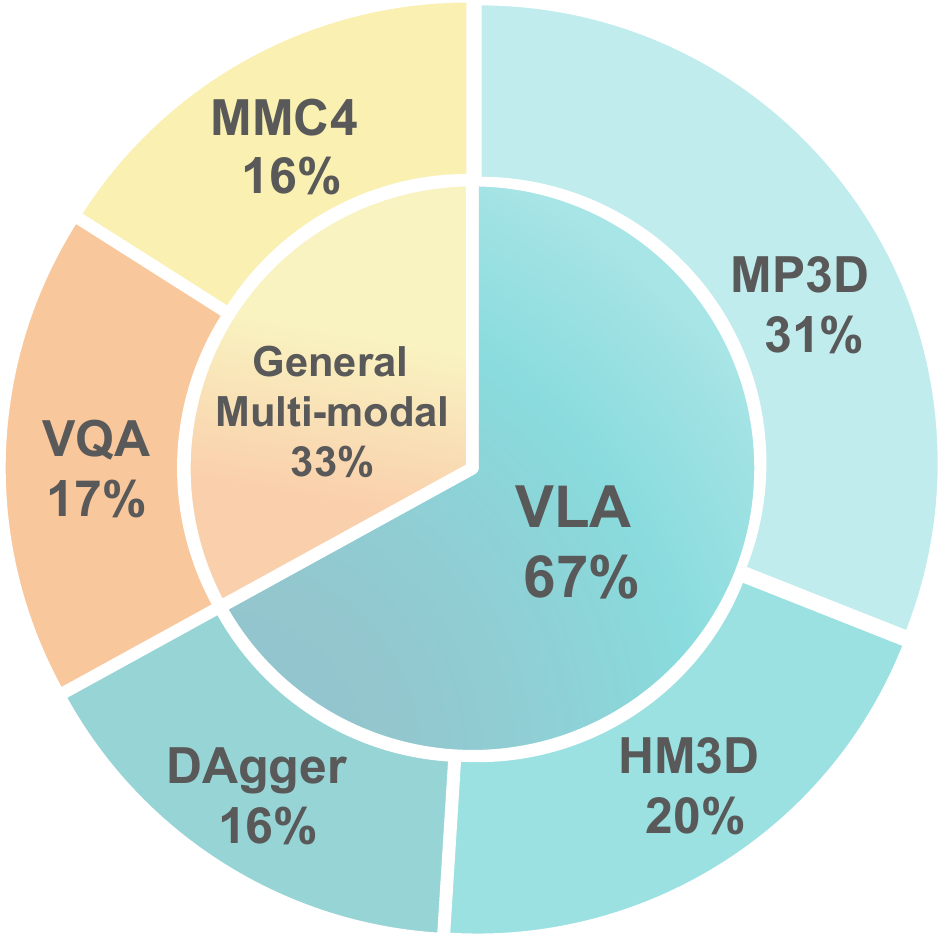}
    \caption{Co-Training Data Recipe of StreamVLN}
    \label{fig:data}
\end{figure}
\noindent\textbf{Vision-Language Action Data.} 
We collect navigation-specific training data using the Habitat simulator across multiple public VLN datasets. Specifically, we collect 450K samples (video clips) from 60 Matterport3D~\cite{Matterport3D} (MP3D) environments, sourced from R2R~\cite{anderson2018r2r}, R2R-EnvDrop~\cite{tan2019envdrop} and RxR~\cite{ku2020rxr}. To further improve generalization through increased scene diversity, we incorporate an additional 300K samples from a subset of ScaleVLN~\cite{wang2023scaling}, spanning 700 Habitat Matterport3D~\cite{ramakrishnan2021habitat} (HM3D) scenes. 
In addition, we adopt the DAgger algorithm to enhance the model's robustness and generalization ability in novel scenes and during error recovery.
Using Habitat's shortest-path follower as the expert policy, we collect corrective demonstrations on model rollouts after the initial training stage.
These DAgger-collected samples (240K) are then incorporated into the training set for co-training.
\begin{table*}[th]
\centering
\setlength{\tabcolsep}{8pt}  
\renewcommand{\arraystretch}{1.4} 
\caption{Comparison with state-of-the-art methods on VLN-CE R2R and RxR Val-Unseen split.}
\begin{tabular}{l|cccc|cccc|cccc}
\toprule
\multirow{2}{*}{Method} & \multicolumn{4}{c|}{Observation} & \multicolumn{4}{c|}{R2R Val-Unseen} & \multicolumn{4}{c}{RxR Val-Unseen} \\ 
\cmidrule(lr){2-5} \cmidrule(lr){6-9} \cmidrule(lr){10-13}
      & Pano. & Odo. & Depth & S.RGB & NE$\downarrow$ & OS$\uparrow$ & SR$\uparrow$ & SPL$\uparrow$ & NE$\downarrow$ & SR$\uparrow$ & SPL$\uparrow$ & nDTW$\uparrow$ \\ 
\midrule
\rowcolor{LightGray}
HPN+DN$^*$~\cite{krantz2021waypoint} & $\checkmark$ & $\checkmark$ & $\checkmark$ &  & 6.31  & 40.0  & 36.0  & 34.0  & - & - & - & - \\
\rowcolor{LightGray}
CMA$^*$~\cite{hong2022bridging}      & $\checkmark$ & $\checkmark$ & $\checkmark$ &  & 6.20  & 52.0  & 41.0  & 36.0  & 8.76 & 26.5 & 22.1 & 47.0 \\
\rowcolor{LightGray}
\vlnbert$^*$~\cite{hong2022bridging} & $\checkmark$ & $\checkmark$ & $\checkmark$ &  & 5.74  & 53.0  & 44.0  & 39.0  & 8.98 & 27.0 & 22.6 & 46.7 \\
\rowcolor{LightGray}
Sim2Sim$^*$~\cite{krantz2022sim}    & $\checkmark$ & $\checkmark$ & $\checkmark$ &  & 6.07  & 52.0  & 43.0  & 36.0  & - & - & - & - \\
\rowcolor{LightGray}
GridMM$^*$~\cite{wang2023gridmm}    & $\checkmark$ & $\checkmark$ & $\checkmark$ &  & 5.11  & 61.0  & 49.0  & 41.0  & - & - & - & - \\
\rowcolor{LightGray}
ETPNav$^*$~\cite{an2023etpnav}      & $\checkmark$ & $\checkmark$ & $\checkmark$ &  & 4.71  & 65.0  & 57.0  & 49.0  & 5.64 & 54.7 & 44.8 & 61.9 \\ 
\rowcolor{LightGray}
ScaleVLN$^{*}$~\cite{wang2023scaling} & $\checkmark$ & $\checkmark$ & $\checkmark$ &  & 4.80  & --    & 55.0  & 51.0  & -& - & - & -\\
\midrule
\rowcolor{LightBlue}
InstructNav~\cite{long2024instructnav} & \checkmark & \checkmark & \checkmark & \checkmark & 6.89 & --   & 31.0 & 24.0 & - & - & - & - \\
\rowcolor{LightBlue}
AG-CMTP~\cite{chen2021topological}   & $\checkmark$ & $\checkmark$ & $\checkmark$ &  & 7.90  & 39.2  & 23.1  & 19.1  & - & - & - & - \\
\rowcolor{LightBlue}
R2R-CMTP~\cite{chen2021topological}  & $\checkmark$ & $\checkmark$ & $\checkmark$ &  & 7.90  & 38.0  & 26.4  & 22.7  & - & - & - & - \\
\rowcolor{LightBlue}
LAW~\cite{raychaudhuri2021law}       &  & $\checkmark$ & $\checkmark$ & $\checkmark$ & 6.83  & 44.0  & 35.0  & 31.0  & 10.90 & 8.0 & 8.0 & 38.0 \\
\rowcolor{LightBlue}
CM2~\cite{georgakis2022cross}        &  & $\checkmark$ & $\checkmark$ & $\checkmark$ & 7.02  & 41.5  & 34.3  & 27.6  & - & - & - & - \\
\rowcolor{LightBlue}
WS-MGMap~\cite{chen2022weakly}       &  & $\checkmark$ & $\checkmark$ & $\checkmark$ & 6.28  & 47.6  & 38.9  & 34.3  & - & - & - & - \\
\rowcolor{LightBlue}
ETPNav + FF~\cite{wang2024sim}  &  & $\checkmark$ & $\checkmark$ & $\checkmark$ & 5.95  & 55.8  & 44.9  & 30.4  & 8.79 & 25.5 & 18.1 & - \\
\rowcolor{LightBlue}
Seq2Seq~\cite{krantz_vlnce_2020}    &  &  & $\checkmark$ & $\checkmark$ & 7.77  & 37.0  & 25.0  & 22.0  & 12.10 & 13.9 & 11.9 & 30.8 \\
\rowcolor{LightBlue}
CMA~\cite{krantz_vlnce_2020}        &  &  & $\checkmark$ & $\checkmark$ & 7.37  & 40.0  & 32.0  & 30.0  & - & - & - & - \\
\midrule

NaVid~\cite{zhang2024navid}   &  &  &  & $\checkmark$ & 5.47  & 49.1  & 37.4  & 35.9  & - & - & - & - \\

MapNav~\cite{zhang2025mapnav}    &  &  &  & $\checkmark$ & \textbf{4.93}  & 53.0  & 39.7  & 37.2  & - & - & - & - \\

NaVILA~\cite{cheng2024navila}   &  &  &  & $\checkmark$ & 5.37  & 57.6  & 49.7  & 45.5  & - & - & - & - \\

\textbf{StreamVLN}        &  &  &  & $\checkmark$ & 5.43  & \textbf{62.5}  & \textbf{52.8}  & \textbf{47.2}  & 6.72 & 48.6 & 42.5 & 60.2 \\
\midrule

NaVILA${\dagger}$~\cite{cheng2024navila}   &  &  &  & $\checkmark$ & 5.22  & 62.5  & 54.0  & 49.0  & 6.77 & 49.3 & 44.0 & 58.8 \\

UniNaVid${\dagger}$~\cite{zhang2024uninavid}   &  &  &  & $\checkmark$ & 5.58  & 53.3  & 47.0  & 42.7  & 6.24 & 48.7 & 40.9 & - \\

\textbf{StreamVLN}${\dagger}$   &  &  &  & $\checkmark$ & 4.90  & 63.6  & 56.4  & 50.2  & \textbf{5.65} & \textbf{54.4} & \textbf{45.4} & \textbf{63.7} \\
\midrule

\textbf{StreamVLN-Prune-Mem}${\dagger}$   &  & $\checkmark^*$ & $\checkmark^*$ & $\checkmark$ & \textbf{4.73}  & \textbf{65.5}  & \textbf{57.4}  & \textbf{51.1}  & 5.72 & 53.9 & 45.1 & 63.3 \\
\textbf{StreamVLN-Prune-All}${\dagger}$   &  & $\checkmark^*$ & $\checkmark^*$ & $\checkmark$ & 4.82  & 65.7  & 56.0  & 48.5  & 5.86 & 53.3 & 44.0 & 63.5 \\
\bottomrule
\end{tabular}
\begin{tablenotes}
\item $*$ indicates methods using the waypoint predictor from~\cite{hong2022bridging}. $\dagger$ denotes methods using additional VLN data beyond the R2R-CE and RxR-CE benchmarks. $\checkmark^*$ indicates that Odo. and Depth are only used for back-projection in the \textbf{training-free voxel pruning.}
\end{tablenotes}
\label{tab:comp-vlnce}
\end{table*}

\noindent\textbf{General Vision-Language Data.}
To retain the general reasoning capabilities of the pretrained Video-LLM, we incorporate a diverse set of multimodal training data that complements navigation supervision. Specifically, we include 248K video-based visual question-answering (VQA) samples sourced from publicly available datasets LLaVA-Video-178K~\cite{zhang2024videoinstructiontuningsynthetic} and ScanQA~\cite{azuma2022scanqa}, which combine general video QA with 3D scene understanding to support spatial-temporal and geometric reasoning. To further augment the model’s capacity for multi-turn vision-language interactions, we incorporate 230K interleaved image-text samples from MMC4~\cite{zhu2023multimodal}, which strengthens its ability to parse and generate contextually coherent responses with interleaved visual and textual reasoning.

\section{Experiments}
\subsection{Experimental Setup}
\noindent\textbf{Simulation Benchmark Setup.}
We evaluate our method on two public VLN-CE~\cite{krantz_vlnce_2020} benchmarks collected from Matterport3D scenes using the Habitat simulator: R2R-CE~\cite{anderson2018r2r} and RxR-CE~\cite{ku2020rxr}. R2R-CE provides 5.6K English trajectories with an average length of 10 meters, while RxR-CE includes 126K multilingual instructions (English, Hindi, Telugu) and features longer, more diverse paths (avg.15 meters). The camera HFOVs are 79° for R2R-CE and RxR-CE. Both benchmarks require realistic indoor navigation under continuous control. As our goal is to assess the generalization ability, we focus on the validation unseen splits of both benchmarks
We report standard VLN metrics, including Navigation Error (NE), Success Rate (SR), Oracle Success Rate (OS), and Success weighted by Path Length (SPL), following prior works.

\noindent\textbf{Real-World Evaluation Setup.}
We perform real world experiments based on a Unitree Go2 robotic dog. The robot is equipped with a upward facing camera (Intel\textsuperscript{{\textregistered}} RealSense\textsuperscript{\texttrademark}  D455) for RGB-D observations. We deploy StreamVLN on a remote workstation with an RTX 4090 GPU. The Go2 robot continuously streams visual data to the 4090 server for inference, which returns executable action commands to the robot. The averge inference (0.27s for 4 actions) and communication (0.2s for indoor and 1.0s for outdoor environments) latency enable real-time physical deployment.
\subsection{Implementation Details}
We build StreamVLN based on the LLaVA-Video~\cite{zhang2024llavavideo} 7B model, which uses Qwen2-7B~\cite{yang2024qwen2} as the language model.
Training is conducted in two stages. First, we fine-tune it for one epoch solely on oracle VLN trajectories. Then, we use the model to collect DAgger trajectories and continue training for an additional epoch with a mixture of VLN and general multimodal data. During the warm-up phase, we apply a peak learning rate of 2e-5 for the language model and 5e-6 for the vision encoder. Each training step processes 128 video clips. Training is in around 1500 A100 GPU hours.
\begin{table}[t]
\centering
\setlength{\tabcolsep}{1.5mm} 
\renewcommand{\arraystretch}{1.4} 
\caption{Comparison on ScanQA Val set.}
\begin{tabular}{l|ccccc}
\toprule
\multirow{2}{*}{{Method}} & \multicolumn{5}{c}{{ScanQA}} \\
  & {Bleu-4} $\uparrow$ & {Rouge} $\uparrow$ & {Meteor} $\uparrow$ & {Cider} $\uparrow$ & {EM} $\uparrow$ \\
\midrule
\rowcolor{LightGray} 
ScanRefer~\cite{yu2019deep} & 7.9 & 30.0 & 55.4 & 11.5 & 18.6 \\
\rowcolor{LightGray}
ScanQA~\cite{azuma2022scanqa}     & 10.1 & 33.3 & 64.9 & 13.1 & 21.0 \\
\rowcolor{LightGray}
3D-VisTA~\cite{zhu20233dvista}        & 10.4 & 35.7 & 69.6 & 13.9 & 22.4 \\
\midrule 
3D-LLM*~\cite{hong20233d} & 12.0 & 35.7 & 69.4 & 14.5 & 20.5 \\
LEO~\cite{huang2023leo}           & 13.2 & 49.2 & 101.4 & 20.0 & 24.5 \\
ChatScene*~\cite{huang2023chatscene} & 14.0 & - & 87.6 & - & - \\
Scene-LLM*~\cite{fu2024scenellm}  & {12.0} & 40.0 & 80.0 & 16.6 & 27.2 \\
\midrule
NaviLLM~\cite{zheng2024navillm}   & 12.0 & 38.4 & 75.9 & 15.4 & 23.0 \\
{NaVILA}~\cite{cheng2024navila} (16)   & 15.2 & 48.3  & 99.8 & 19.6  & 27.4 \\
\textbf{StreamVLN} (16) & \textbf{15.7} & \textbf{48.3} & \textbf{100.2} & \textbf{19.8} &\textbf{28.8} \\
\bottomrule
\end{tabular}
\begin{tablenotes}
\item $*$ indicates task-specific fine-tuning. (16) means using 16 frames.
\end{tablenotes}
\label{tab:comp-scanqa}
\end{table}

\begin{figure}
    \includegraphics[width=\linewidth]{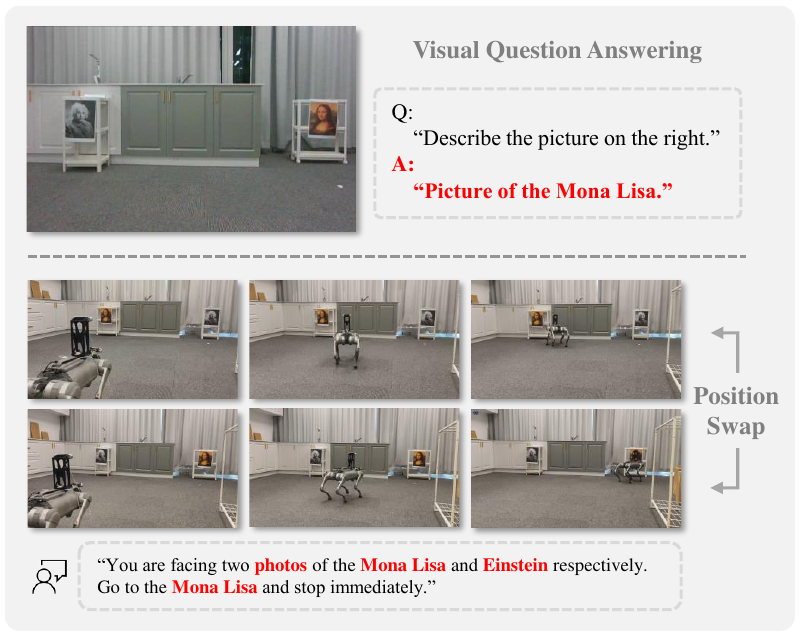}
    \caption{StreamVLN transfers visual reasoning ability to interpreting out-of-domain navigation instructions.}
    \label{fig:choice}
\end{figure}
\begin{figure*}
    \centering
    \includegraphics[width=1.0\linewidth]{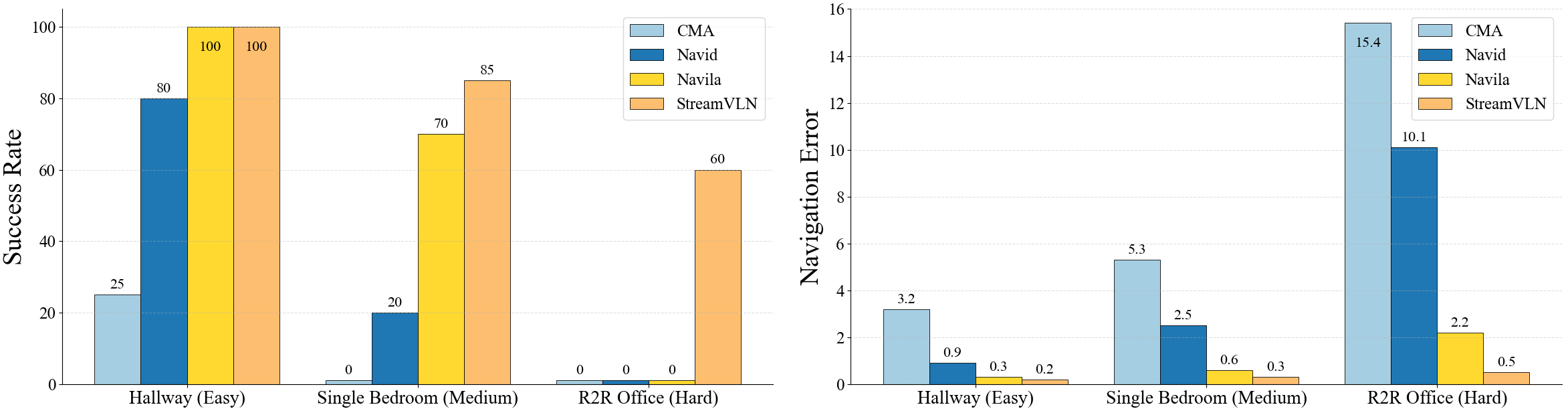}
    \caption{Real-world experiments conducted across hallway (easy), bedroom (medium, single-room), and office (hard, room-to-room) scenarios.}
    \label{fig:real_world_metric}
\end{figure*}
\subsection{Comparisons with State-of-the-Arts}
\noindent\textbf{Results on VLN-CE benchmark.}
Table~\ref{tab:comp-vlnce} shows the performance of our method on the VLN-CE R2R and RxR benchmarks under the Val-Unseen setting, compared with existing VLN-CE methods.  StreamVLN achieves state-of-the-art performance among RGB-only methods both without and with extra navigation datasets, reaching $56.9\%$ SR and $51.9\%$ SPL on R2R, and $52.9\%$ SR and $46.0\%$ SPL on RxR. These results highlight the robustness of StreamVLN across both standard and long-horizon navigation tasks.
Notably, StreamVLN performs comparably to ETPNav~\cite{an2023etpnav}, despite not relying on additional panoramic or waypoint supervision. Furthermore, compared to HMAT trained on entire ScaleVLN dataset with 3 million trajectories, StreamVLN surpasses it with a small subset of ScaleVLN (150k), showing better data efficiency.\\
\noindent\textbf{Effectiveness of Voxel-Based Spatial Pruning}.
Table~\ref{tab:comp-vlnce} bottom shows the effect of applying voxel-based spatial pruning during inference.
We further evaluate two settings: 1) \underline{StreamVLN-Prune-Mem}. Pruning is applied only on 8 memory frames, reducing memory tokens by 28\% on R2R and 22\% on RxR without noticeably affecting performance. 
On the shorter-horizon R2R benchmark, performance even improves, indicating that the current memory still contains redundant tokens for some tasks. Proper pruning thus helps the model focus on relevant tokens, thereby enhancing navigation accuracy. \\
2) \underline{StreamVLN-Prune-All}. Pruning is applied to both memory tokens and the KV cache of streaming frame tokens, reducing visual tokens by 32\% on R2R and 30\% on RxR, with only a slight decrease in performance.

\noindent\textbf{Results on Video Question Answering.}
To evaluate StreamVLN’s spatial scene understanding capabilities, we conduct experiments on the widely-used ScanQA benchmark for 3D question answering based on real-world scans. StreamVLN answers questions by analyzing 16 multi-view images from each scan. As shown in Table~\ref{tab:comp-scanqa}, StreamVLN outperforms state-of-the-art generalist navigation models such as NaviLLM~\cite{zheng2024navillm} and NaVILA~\cite{cheng2024navila}. As shown in Figure~\ref{fig:choice}, we observe that the strong VQA capabilities contribute to better generalization to novel navigation instructions.
\begin{figure*}
    \centering
    \includegraphics[width=1.0\linewidth]{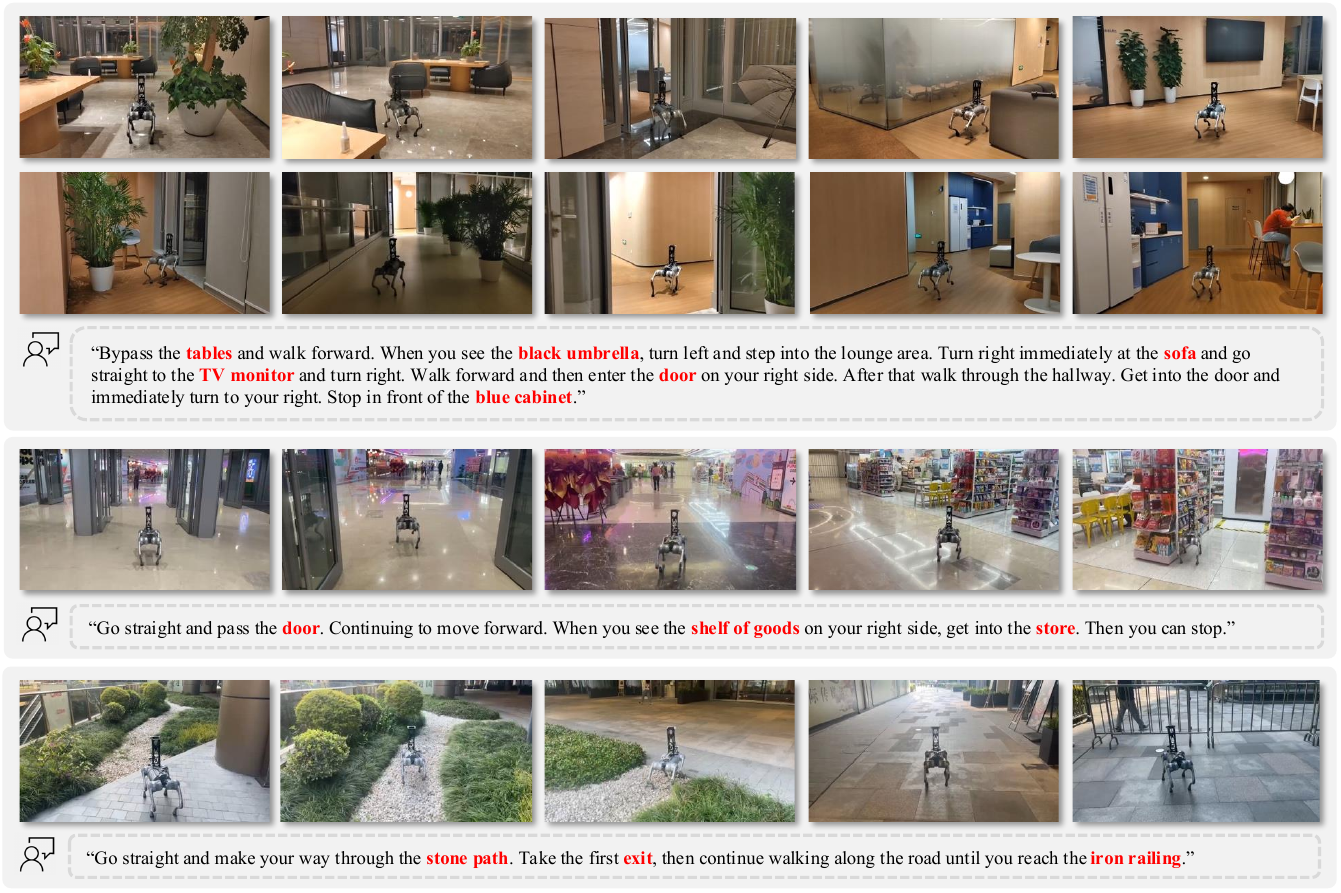}
    \caption{Qualitative results of StreamVLN in several representative real-world environments. From top to bottom are \texttt{Home}, \texttt{Workspace}, \texttt{Mall} and \texttt{Outdoor}. 
    StreamVLN achieves robust performance across diverse VLN scenarios, capable of accurately following complex instructions with various landmarks (marked as \textcolor{red}{red}) and handling real-world disturbances.}
    \label{fig:realworld}
    \vspace{-4mm}
\end{figure*}

\noindent\textbf{Real-World Experimental Results.} 
To quantitatively assess StreamVLN in real-world settings, we evaluated its performance in three distinct scenarios: hallway (easy), bedroom (medium difficulty, single-room), and office (hard, room-to-room). The baselines include the traditional learning-based CMA~\cite{hong2022bridging} and VLM-based approaches such as NaVid~\cite{zhang2024navid} and NaVILA~\cite{cheng2024navila}. For each method, we conducted 20 trials per scenario, measuring Success Rate (SR) and Navigation Errors (NE). As shown in Figure~\ref{fig:real_world_metric}, CMA performs poorly, while NaVid succeeds only on simple, short-horizon tasks. NaVILA can follow longer-horizon instructions but fails in complex office scenarios. In contrast, StreamVLN successfully completes both simple and challenging long-horizon navigation tasks.
Qualitative results in \texttt{Workspace}, \texttt{Mall}, and \texttt{Outdoor} are presented in Figure~\ref{fig:realworld}. 
Especially, the success cases in \texttt{Mall} and \texttt{Outdoor} environments highlight StreamVLN’s strong generalization to novel scenes and tasks.
Please refer to the \textcolor{blue}{\textit{demo video}} for full demonstrations.
\subsection{Ablation Studies}
\noindent\textbf{Data Ablation.}
\begin{table}[t]
\centering
\setlength{\tabcolsep}{0.8mm}
\renewcommand{\arraystretch}{1.5}
\caption{
Ablation study of different training data compositions on VLN-CE R2R Val-Unseen split.
}
\begin{tabular}{ccccc|cccc}
\toprule
R2R & RxR & DAgger & VL Data & ScaleVLN & NE$\downarrow$ & OS$\uparrow$ & SR$\uparrow$ & SPL$\uparrow$ \\
\midrule
\rowcolor{LightGray}
\checkmark & \checkmark &  &  &  & 5.98 & 51.3 & 45.6 & 42.3 \\
\checkmark & \checkmark & \checkmark & VidQA &  & 5.47 & 57.8 & 50.8 & 45.7 \\
\checkmark & \checkmark & \checkmark & VidQA+MMC4 &  & 5.43 & 62.5 & 52.8 & 47.2 \\
\checkmark & \checkmark & \checkmark & VidQA+MMC4 & \checkmark & \textbf{4.90} & \textbf{63.6} & \textbf{56.4} & \textbf{50.2} \\
\checkmark & \checkmark &  & VidQA+MMC4 & \checkmark & 5.73 & 56.4 & 50.2 & 47.1 \\
\checkmark &  & \checkmark & VidQA+MMC4 & \checkmark & 5.90 & 55.9 & 47.9 & 43.6 \\
\bottomrule
\end{tabular}
\label{tab:ablationdataset}
\end{table}

Table~\ref{tab:ablationdataset} presents an ablation study on different training data compositions. All results are reported without using voxel-based spatial pruning.
The first row shows the first-stage performance when training with only oracle navigation data.
After collecting DAgger data, we co-train oracle data, DAgger data, and vision-language (VL) data. 
In the second row, we use only VideoQA data as VL data. 
While the third row mixes VideoQA and MMC4 (M) data in an interleaved image-text format.
For a fair comparison, the total number of VL Data is kept the same. 
We can observe that the second-stage co-training brings significant gains (+5.3 SR / +4.1 SPL) and incorporating MMC4 further improves performance (+2.0 SR / +1.5 SPL).
Comparing the third and fourth rows, we see that adding ScaleVLN data brings additional gains (+2.9 SR / +3.7 SPL).
To assess the importance of DAgger data, we remove it from the co-training data, as shown in the fifth row. The results show that DAgger data plays a crucial role in boosting performance (+5.5 SR / +3.8 SPL). Furthermore, the last row highlights that incorporating RxR data yields notable performance gains (+7.8 SR / +7.3 SPL).
\begin{table}
\centering
\setlength{\tabcolsep}{3.5mm}
\renewcommand{\arraystretch}{1.5} 
\caption{Ablation on the impact of different memory context sizes and sliding window sizes on VLN-CE R2R Val-Unseen split.
}
\begin{tabular}{cc|cccc}
\toprule
 Memory & Window & NE$\downarrow$ & OS$\uparrow$ & SR$\uparrow$ & SPL$\uparrow$ \\
\midrule
 2*196 & 8 & 6.96 & 48.2 & 37.3 & 34.2 \\
 4*196 & 8 & 6.62 & 49.1 & 38.9 & 35.4 \\
 8*196 & 8 & \textbf{6.05} & \textbf{53.8} & \textbf{45.5} & \textbf{41.6} \\
 \textit{all}   & 8 & 6.76 & 49.5 & 40.0 & 36.4 \\
 8*196 & 4 & 6.31 & 51.1 & 41.4 & 37.5 \\
 8*196 & 2 & 6.16 & 52.8 & 43.7 & 40.3 \\
\bottomrule
\end{tabular}
\label{tab:long_term}
\end{table}
\begin{figure}
    \centering
    \includegraphics[width=0.85\linewidth]{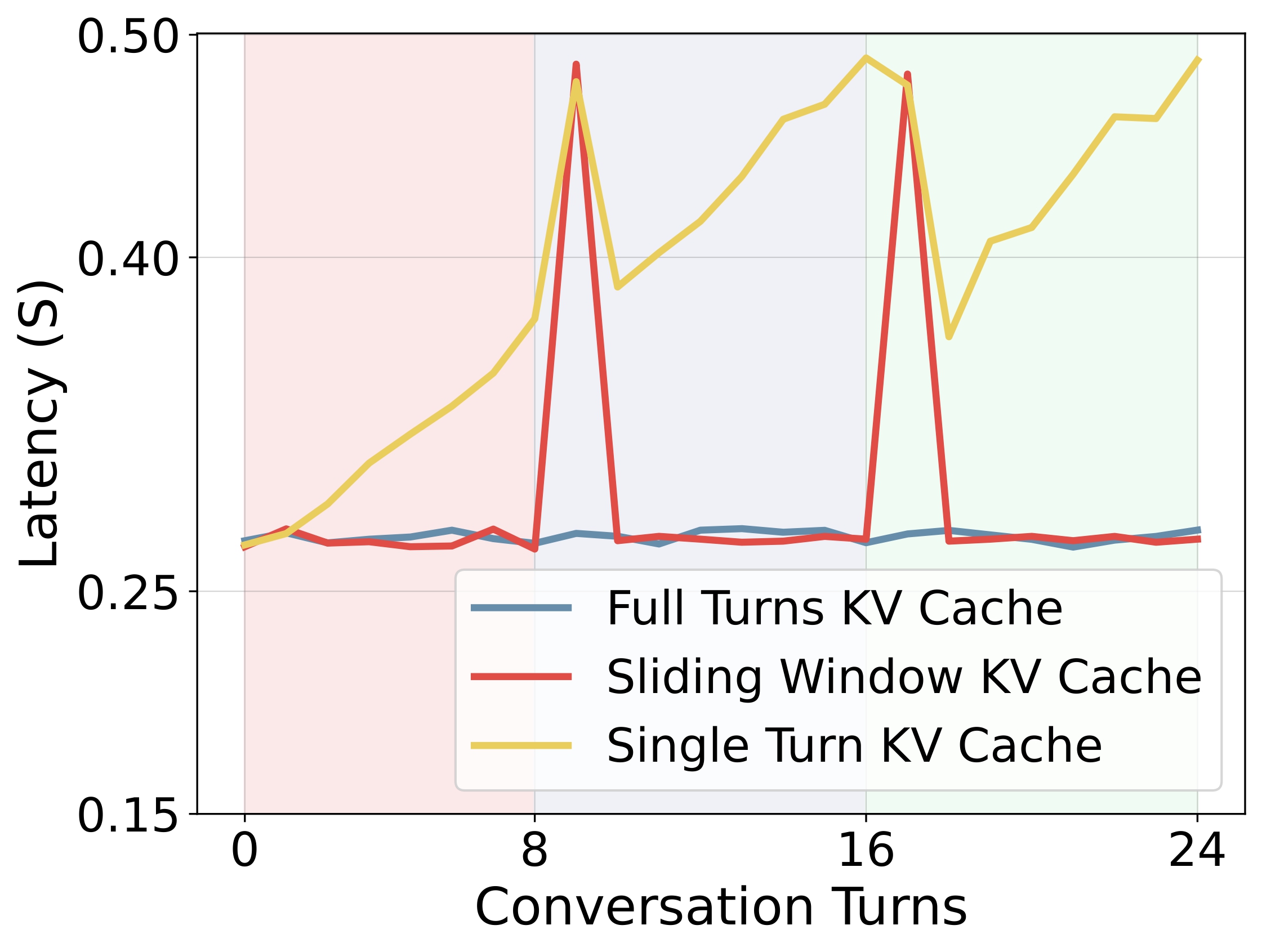}
    \caption{Impact of KV cache reuse across multiple turns . The Sliding window size is 8 (8 conversation turns) in all three settings.}
    \label{fig:kv_cache}
\end{figure}

\noindent\textbf{Memory Context Size.}
We study the impact of the memory context size in the hybrid context modeling strategy. 
As shown in Table~\ref{tab:long_term} (results shown are from first-stage training using only oracle VLN data), increasing the memory size from $2*196$ to $8*196$ while keeping the window size fixed at 8 significantly improves navigation performance, with SR rising from 37.3 to 45.5. This indicates the importance of fine-grained memory in supporting long-horizon reasoning. Notably, using the entire visual context (\textit{all}) as memory doesn't yield the best results, suggesting that an overly long and varied context token sequence may introduce bias in training and hinder generalization at test time.

\noindent\textbf{Sliding Window Size.}
We also evaluate the effect of the number of dialogue turns retained in the sliding window in Table~\ref{tab:long_term}. A smaller window size leads to more frequent shifts, resulting in a significantly larger number of training samples. For example, a window size of 8 yields approximately 450K samples, while sizes of 4 and 2 increase this to 815K and 1.5M respectively. This growth not only raises the training cost linearly but also introduces greater class imbalance, which may affect training stability. We find that retaining 8 continuous dialogue turns achieves the best balance—delivering strong navigation performance while maintaining the lowest training cost.

\noindent\textbf{Effectiveness of KV-Cache Reuse.}
We evaluate the impact of KV cache reuse on the decoding latency under different settings.
As shown in Figure~\ref{fig:kv_cache}, reusing the KV cache across all dialogue turns (Full Turns) achieves consistently low latency—since only the current observation tokens require prefill computation for generating the 8 action tokens—\textit{but storing all the cache poses significant memory overhead}.
If the KV cache is maintained only within 8 turns (Sliding Window), the decoding latency will increase at the beginning of each sliding window due to the need to prefill the previous window context tokens.
Under the Single Turn setting, where the KV cache is not reused across turns (as in prior work), decoding latency steadily increases with the number of turns. Turns 0–8 incur lower latency since no historical context is included, while turns 8–16 and 16–24 have similar latency growth with a fixed memory size.

\section{Conclusion}
This paper presents \textit{StreamVLN}, a new streaming vision-language-navigation framework based on Video-LLMs.
Compared to previous Video-LLM-based VLN methods that treat each interaction as an independent dialogue and refresh history at every step, StreamVLN can reuse past key/value (KV) states through a hybrid memory design.
By maintaining a fast-updating sliding window for immediate responsiveness and a slow-updating long-term memory for temporal reasoning, StreamVLN enables efficient, coherent, and scalable action generation over long video streams.
Empirical results on standard VLN-CE benchmarks demonstrate that StreamVLN achieves superior performance with lower latency, paving the way for real-time long-horizon navigation.

\noindent \textbf{Acknowledgements.} This work is supported by Shanghai Artificial Intelligence Laboratory. The research work described in this paper was conducted in the JC STEM Lab of Autonomous Intelligent Systems funded by The Hong Kong Jockey Club Charities Trust.
\bibliographystyle{IEEEtran}
\bibliography{IEEEabrv,root}

\end{document}